\documentclass[11pt,a4paper]{article}
\usepackage[hyperref]{AACL-IJCNLP2020}

\usepackage{times}
\usepackage{latexsym}
\usepackage{booktabs}
\usepackage{tabulary}
\usepackage{amsmath}

\usepackage{microtype}
\usepackage{linguex}
\usepackage{array,multirow,graphicx}
\aclfinalcopy % Uncomment this line for the final submission
\def\aclpaperid{314} %  Enter the acl Paper ID here

\newcommand{\mytilde}{\raise.17ex\hbox{$\scriptstyle\mathtt{\sim}$}}
\newcommand{\baseline}{\textsc{Base}}
\newcommand{\sent}{\textsc{Sim}}
\newcommand{\translate}{\textsc{Translate}}

\title{Asking Crowdworkers to Write Entailment Examples: \\The Best of Bad Options}

\author{Clara Vania \quad Ruijie Chen \quad Samuel R. Bowman \\
  New York University \\
  {\tt \{c.vania, rc3959, bowman\}@nyu.edu}}
  
\date{}

\begin{document}
\maketitle
\begin{abstract}
Large-scale natural language inference (NLI) datasets such as SNLI or MNLI have been created by asking crowdworkers to read a \textit{premise} and write three new \textit{hypotheses}, one for each possible semantic relationships (\textit{entailment}, \textit{contradiction}, and \textit{neutral}).
While this protocol has been used to create useful benchmark data, it remains unclear whether the writing-based annotation protocol is optimal for any purpose, since it has not been evaluated directly. 
Furthermore, there is ample evidence that crowdworker writing can introduce artifacts in the data.
We investigate two alternative protocols which automatically create candidate (\textit{premise, hypothesis}) pairs for annotators to label.
Using these protocols and a writing-based baseline, we collect several new English NLI datasets of over 3k examples each, each using a fixed amount of annotator time, but a varying number of examples to fit that time budget.
Our experiments on NLI and transfer learning show negative results: None of the alternative protocols outperforms the baseline in evaluations of generalization within NLI or on transfer to outside target tasks. 
We conclude that crowdworker writing still the best known option for entailment data, highlighting the need for further data collection work to focus on improving \textit{writing-based} annotation processes.

\end{abstract}

\section{Introduction}
\label{sec:intro}

Research on natural language understanding has benefited greatly from the availability of large-scale, annotated data, especially for tasks like reading comprehension and natural language inference, which lend themselves to non-expert crowdsourcing. These datasets are useful in three settings: evaluation \citep{williams-etal-2018-broad,rajpurkar-etal-2018-know,zellers-etal-2019-hellaswag}; pretraining \citep{phang2018sentence,conneau-etal-2018-xnli,pruksachatkun2020intermediatee}; and as training data for downstream tasks \citep{trivedi-etal-2019-repurposing,portelli-etal-2020-distilling}.

Natural language inference (NLI), also known as \textit{recognizing textual entailment} \citep[RTE;][]{rte} is the problem of determining whether or not a hypothesis semantically entails a premise. The two largest NLI corpora, SNLI \citep{bowman-etal-2015-large} and MNLI \citep{williams-etal-2018-broad} are created by asking crowdworkers to write three labeled \textit{hypothesis} sentences given a \textit{premise} sentence taken from a preexisting text corpus. While these datasets have been widely used as benchmarks for NLU, there have been no studies evaluating writing-based annotation for collecting NLI data. Moreover, there is growing evidence that human writing can introduce \textit{annotation artifacts}, which enable models to perform moderately well just by learning spurious statistical patterns in the data \citep{gururangan-etal-2018-annotation,tsuchiya-2018-performance,poliak-etal-2018-collecting}.

\begin{figure*}[ht]
    \centering
    \includegraphics[width=.95\linewidth]{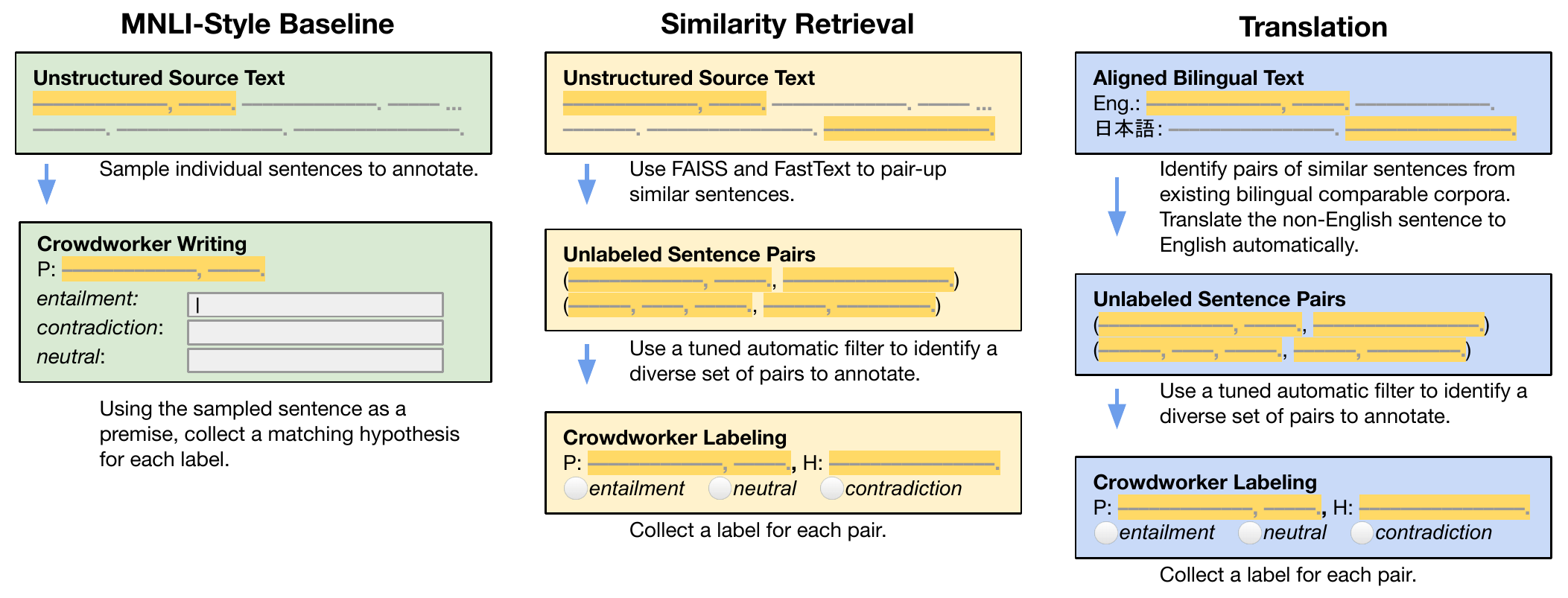}
    \caption{We introduce two new protocols for natural language inference data collection. Both use fully-automated pipelines to generate pairs of semantically-related sentences, which crowdworker annotators then label.}
    \label{fig:protocols}
\end{figure*}

This paper explores the possibility of collecting high-quality NLI data without asking crowdworkers to write hypotheses. We introduce two alternative protocols (Figure~\ref{fig:protocols}) which substitute crowdworker writing with fully-automated pipelines to generate premise-hypothesis sentence pairs, which annotators then simply label. The first protocol uses a sentence-similarity-based method to pair similar sentences from large unannotated corpora. The second protocol uses parallel sentences and uses machine translation systems to generate sentence pairs. Using the MNLI protocol as our baseline, we collect five datasets using premises taken from Gigaword news text \citep{Gigaword} and Wikipedia. We then compare models trained using these datasets for their generalization performance within NLI and for transfer learning to other tasks.

We start from the assumption that writing a new hypothesis takes more time and effort than simply labeling a presented hypothesis. As a result, it is plausible that our protocols could offer some value even if the quality of the data they produce is no better than a writing-based baseline. To study the cost trade-off, we collect each dataset under the same fixed annotation budget with a fixed ($\sim$ US \$15) hourly wage. Using this constraint, we collect approximately twice as many examples from our new protocols.

Our main results on natural language inference and transfer learning are clearly negative. Human-constructed examples appear to be far superior to automatically-constructed examples in both settings. While crowdworker writing in data collection has known issues, it produces better training data than our automatic methods, or any known comparable methods which intervene the writing-based protocol to help crowdworkers with the writing process \citep{bowman2020collecting}.
This strongly suggests that future work on data quality should focus on improving human-based generation processes.

\begin{table*}[ht]
    \small
    \setlength{\tabcolsep}{.3em}
    \tymin=20pt
    \tymax=200pt
    \begin{tabulary}{1.0\textwidth}{llLL}
    \toprule
    \textbf{Dataset} & \textbf{Label} & \textbf{Premise} & \textbf{Hypothesis} \\
    \midrule
    Base-News & \textbf{E} & The city reconsidered that position on Wednesday, saying it was seeking to raise an additional \$1.5 million to extend Mardi Gras over two weekends and to pay for overtime on several days. & The city is looking to get more money for Mardi Gras. \\
    \addlinespace[.15cm]
    Base-Wiki & \textbf{C} & Service books were not included and a note at the end mentions many other books in French, English and Latin which were then considered worthless. & Service books were included. \\
    \addlinespace[.15cm]
    Sim-News & \textbf{N} & All of them run out like college football players before a big bowl game. & Pray before a college football game. \\
    \addlinespace[.15cm]
    Sim-Wiki & \textbf{C} & His work was heavily criticised as unscientific by his contemporaries. & His work was recognized and admired by his contemporaries. \\
    \addlinespace[.15cm]
    Translate-Wiki & \textbf{E} & This was used to indicate a positive response, or truth, or approval of the item in front of it. & This was used to indicate yes, true, or confirmed on items in a list. \\
    \bottomrule
    \end{tabulary}
    \caption{Examples of sentence pairs chosen randomly from each test set, along with their assigned labels. \textbf{E}: \textit{entailment}, \textbf{C}: \textit{contradiction}, \textbf{N}: \textit{neutral}.}
    \label{tab:sent-examples}
\end{table*}

\section{Collecting NLI Data}

We compare three protocols for collecting NLI data: (1) a baseline MNLI-style protocol (\textbf{\sc{Base}}), (2) a sentence-similarity-based protocol (\textbf{\sc{Sim}}), and (3) a translation-based protocol (\textbf{\sc{Translate}}). To test generalization performance across domains, we collect two datasets for \textsc{Base} and \textsc{Sim}, using text from Gigaword (news) and Wikipedia (wiki) domains.\footnote{The premise sentences for each protocol can be different although they come from the same source.} For \textsc{Translate}, we collect a dataset from WikiMatrix \citep{wikimatrix}, a collection of Wikipedia parallel sentences. Table~\ref{tab:sent-examples} shows examples of sentence pairs collected using each protocol.

Our new protocols (Figure~\ref{fig:protocols}) share a similar automated pipeline. Given an unstructured text, we automatically collect similar sentence pairs which annotators then label. There are two key differences between our new protocols and \baseline.\ First, our automatically paired sentences are \textit{unlabeled}, and thus require a further data labeling process (Section \ref{sec:labeling}). Second, our protocols might produce datasets with imbalanced label distributions. This is in contrast to \baseline,\ which ensures each premise will have one hypothesis for each label in the annotation. The following subsections describe each protocol in more detail.

\subsection{Baseline (\sc{Base})}

Our \baseline\ protocol closely follows that used for MNLI. We randomly sample premise sentences from Gigaword and Wikipedia and ask crowdworkers to write three new hypotheses, one for each relation type.\footnote{Our instructions can be found in the Appendix~\ref{sec:appendix}, and our FAQs are available at \url{https://sites.google.com/nyu.edu/nlu-mturk-faq/writing-sentences}.}

\subsection{Sentence Similarity (\sc{Sim})}
\label{sec:sentsim}

Our \sent\ protocol exploits the fact that, in large corpora, it should be easy to find pairs of sentences that describe similar events or situations. For example, in Gigaword, one event might be written differently by different news sources in ways that yield any of our three relationships. We collect similar sentences and automatically match them to form sentence pairs which annotators then label. The whole pipeline consists of three steps: indexing and retrieval, reranking, and crowdworker labeling.

\paragraph{Indexing and Retrieval}

Given a raw text, we first split it into sentences.\footnote{We use Spacy's \texttt{"en\_core\_web\_lg"} model to segment sentences and extract noun phrase and entities for later use in reranking.} We encode each sentence as a 300-dimensional vector using fastText \citep{bojanowski-etal-2017-enriching} and index them using FAISS \citep{FAISS}, an open-source library for large-scale similarity search on vectors.\footnote{\url{https://github.com/facebookresearch/faiss}} Since Gigaword and Wikipedia consist of billions of sentences, we perform dimensionality reduction using PCA and cluster the search space to allow efficient index and retrieval. We randomly sample query sentences from the text corpus and retrieve the top 1k most similar sentences for each query. This is done by building an index with type \texttt{"PCAR64,IVFx,Flat"} in FAISS terms, where \texttt{x} varies depending on the corpus size. Details of our indexing and retrieval procedures can be found in Appendix~\ref{app:indexing}.

\paragraph{Reranking}

FastText uses a Continuous Bag-of-Words (CBoW) model to learn word representations. This means given a query, we will sometimes have top matches which are syntactically similar but describe different events or situations. While unrelated sentences can be contradictory or neutral, directly using the top-$n$ sentences from FAISS will give us too few entailment pairs. Furthermore, because we use randomly sampled sentences as queries, there could be no good match at all for a given query. 

To collect a set of sentence pairs with a reasonable label distribution, for each query, we retrieve top-$K$ matches and rerank the (query, retrieved sentence) pairs using the following features:
\begin{itemize}
    \item \textit{FAISS similarity score}: The raw similarity score from FAISS.
    \item \textit{Word types}: The proportion of word types in the query sentence seen in the retrieved sentence.
    \item \textit{Noun phrase}: The proportion of noun phrases in the query sentence seen in the retrieved sentence.
    \item \textit{Subjects}: The proportion of complete subject spans (some sentences with embedded clauses can have more than one subject) in the query sentence seen in the retrieved sentence.
    \item \textit{Named entity}: The proportion of named entities in the query sentence seen in the retrieved sentence.
    \item \textit{Time}: A boolean feature which denotes whether two sentences are written in the same month and year (only for Gigaword)
    \item \textit{Wiki article}: A boolean feature which denotes whether the pairs come from the same article. (only for Wikipedia)
    \item \textit{Wiki link}: The proportion of hyperlink tokens in the query sentence seen in the retrieved sentence (only for Wikipedia)
\end{itemize}
The choice of these hand-crafted features will likely impact the distribution of our final dataset, but we don't expect these choices to inject significant label-association artifacts, since our methods play no role in setting labels. We calculate the score for each sentence pair using a weighted sum of these features. We populate pairs from all queries and sort them based on their feature scores. We then select the top $N$\% pairs as our final pairs. 

We use a Bayesian hyperparameter optimization to tune the feature weights, $K$, and $N$. In an ideal case, we want our dataset to have a balanced distribution so that all classes will be represented equally. To push for this, we tune these parameters to minimize the Kullback–Leibler (KL) divergence between a uniform distribution across three entailment classes, $P(x)$, and an empirical distribution, $Q(x)$, computed based on the predictions of an NLI model. We run Bayesian optimization for 100 iterations using Optuna \citep{optuna_2019}. For the NLI model, we use a RoBERTa$_{\rm{Large}}$ model fine-tuned on a combination of SNLI, MNLI, and ANLI. 

\subsection{Translation (\sc{Translate})}
\label{sec:sentcomp}

Multilingual comparable corpora contain \textit{similar} texts in at least two different languages. If they are sentence-aligned, we can automatically translate text from one language to one of the others to yield candidate sentence pairs. Since the alignment behind the corpus can be noisy, the resulting sentence pairs range almost continuously from being parallel to being semantically unrelated, potentially fitting any of the three entailment relationships. In the \translate\ protocol, we investigate whether we can use such sentence pairs as entailment data.

We use WikiMatrix \citep{wikimatrix}, a collection of 135 million Wikipedia parallel sentences, which was constructed by aligning similar sentences in different languages in a joint sentence embedding space \citep{schwenk-2018-filtering,artetxe-schwenk-2019-margin}. It is a mix of translated sentence pairs and comparable sentences written independently about the same information. We collect parallel sentences where one of the sentences is in English, $s^E$. For the paired non-English languages, we pick 5 languages: German, French, Indonesian, Japanese, and Czech. We then translate the aligned non-English sentence into an English sentence, $s^{\hat{E}}$ using the OPUS-MT \citep{TiedemannThottingal:EAMT2020} machine translation systems, and treat $(s^E, s^{\hat{E}})$ as a sentence pair. The diverse set of languages allows us to collect a more diverse set of sentence pairs coming from the structural differences across languages. We do not perform any reranking as our predictions using an NLI model on the initially retrieved data (the same one that we used in \textsection{\ref{sec:sentsim}}) shows a near-balanced distribution.

\subsection{Data Labeling}
\label{sec:labeling}
We use Amazon Mechanical Turk to label the automatically-collected sentence pairs (\sent\ and \translate). We hire crowdworkers which have completed at least 5000 HITs with at least a 99\% acceptance rate. In each task, we present crowdworkers with a sentence pair and ask them to provide a single label (\textit{entailment}, \textit{contradiction}, \textit{neutral} or \textit{``I don't understand"}) for the pair. The latter is used if there are problems with either sentence, e.g., because of errors during preprocessing. We collect one label per sentence pair. We use the same HIT setup for validating our test sets (Section \ref{sec:dataset}).

\begin{table}[t]
    \centering
    \small
    \begin{tabular}{lrr}
    \toprule
    & \textbf{Individual == Gold} & \textbf{No Gold Label} \\
    \midrule
    MNLI (Full) & 88.7\% & 1.8\% \\
    \cmidrule{1-3}
    Base-News & 78.7\% & 13.1\% \\
    Base-Wiki & 76.4\% & 10.0\% \\
    Sim-News & 72.9\% & 15.8\% \\
    Sim-Wiki & 74.1\% & 11.9\% \\
    Translate-Wiki & 72.8\% & 14.6\% \\
    \bottomrule
    \end{tabular}
    \caption{Validation statistics for each protocol, compared to MNLI Full.}
    \label{tab:agreement}
\end{table}

\begin{table*}[ht]
    \centering
    \small
    \begin{tabular}{llrrrrrrrrrrrrr}
    \toprule
     & & \textbf{\#Pairs} &
     \multicolumn{3}{c}{\textbf{Label Distribution}} &
     \multicolumn{2}{c}{\textbf{HL$_\text{E}$}} &
     \multicolumn{2}{c}{\textbf{HL$_\text{C}$}} &
     \multicolumn{2}{c}{\textbf{HL$_\text{N}$}} &\multicolumn{3}{c}{\textbf{Word Type Overlap}} \\
     & & & 
     \multicolumn{1}{c}{E} & 
     \multicolumn{1}{c}{C} & 
     \multicolumn{1}{c}{N} & 
     \multicolumn{1}{c}{$\mu$} & 
     \multicolumn{1}{c}{($_\sigma$)} &
     \multicolumn{1}{c}{$\mu$} & 
     \multicolumn{1}{c}{($_\sigma$)} &
     \multicolumn{1}{c}{$\mu$} & 
     \multicolumn{1}{c}{($_\sigma$)} &
     \multicolumn{1}{c}{E} & 
     \multicolumn{1}{c}{C} & 
     \multicolumn{1}{c}{N} \\
     \midrule
     \midrule
    \parbox[t]{2mm}{\multirow{6}{*}{\rotatebox[origin=c]{90}{\textbf{Training}}}}
    & MNLI-3k & 2750 & 33.4 & 33.9 & 32.7 & 9.7 & 4.4 & 9.4 & 4.0 & 11.0 & 4.4 & 25.2 & 17.3 & 15.4 \\
    \cmidrule{2-15}
    & Base-News & 2734 & 33.5 & 33.4 & 33.2 & 12.1 & 6.0 & 11.8 & 5.8 & 12.4 & 6.2 & 23.5 & 18.4 & 18.1 \\
    & Base-Wiki & 2740 & 33.3 & 33.7 & 33.0 & 11.1 & 7.7 & 10.5 & 4.5 & 11.6 & 7.1 & 31.2 & 23.4 & 22.7 \\
    & Sim-News & 6627 & 21.8 & 39.1 & 39.2 & 23.2 & 9.7 & 22.7 & 10.0 & 23.3 & 9.9 & 46.6 & 21.8 & 23.0 \\
    & Sim-Wiki & 6174 & 23.5 & 40.4 & 36.1 & 12.8 & 6.0 & 12.7 & 5.2 & 13.1 & 5.3 & 52.7 & 31.7 & 29.7 \\
    & Translate-Wiki & 6189 & 34.7 & 31.4 & 34.0 & 18.6 & 9.6 & 14.2 & 7.5 & 16.0 & 8.8 & 41.3 & 20.0 & 24.6 \\
    \midrule
    \midrule
    \parbox[t]{2mm}{\multirow{6}{*}{\rotatebox[origin=c]{90}{\textbf{Test}}}}
    & MNLI-3k & 250 & 29.2 & 37.6 & 33.2 & 10.6 & 4.6 & 9.4 & 3.7 & 10.7 & 4.2 & 26.3 & 14.6 & 15.9 \\
    \cmidrule{2-15}
    & Base-News & 226 & 38.1 & 33.2 & 28.8 & 12.8 & 5.7 & 11.5 & 5.1 & 11.6 & 4.6 & 22.8 & 14.4 & 13.5 \\
    & Base-Wiki & 234 & 32.5 & 32.1 & 35.5 & 12.5 & 8.6 & 11.7 & 8.2 & 11.5 & 4.8 & 32.9 & 24.6 & 21.1 \\
    & Sim-News & 219 & 20.1 & 44.3 & 35.6 & 22.5 & 11.1 & 24.9 & 11.1 & 23.9 & 10.9 & 69.3 & 20.9 & 20.6 \\
    & Sim-Wiki & 229 & 20.5 & 45.0 & 34.5 & 12.6 & 7.6 & 13.7 & 5.8 & 12.0 & 4.5 & 60.5 & 32.8 & 28.7 \\
    & Translate-Wiki & 222 & 40.5 & 29.3 & 30.2 & 18.7 & 8.5 & 13.0 & 6.9 & 14.3 & 6.7 & 46.3 & 15.1 & 21.1 \\
    \bottomrule
    \end{tabular}
    \caption{Dataset statistics. \textbf{HL} denotes the \textit{average} and \textit{standard deviation} of the hypothesis length of each label.}
    \label{tab:data-stat}
\end{table*}

\section{The Resulting Datasets}
\label{sec:dataset}

Using \baseline,\ we collect 3k examples for Base-News and Base-Wiki.\footnote{Our preliminary experiments on subsets of MNLI show that RoBERTa performance starts to stabilize once we use at least 3k training examples.} For \sent\ and \translate,\ we increase the number of pairs to exhaust the same budget that was used for the corresponding baseline dataset (\$1,791 for Base-News and \$1,445 for Base-Wiki), allowing us to collect around twice as many examples for each protocol.\footnote{The resulting datasets are available at \url{https://github.com/nyu-mll/semi-automatic-nli}. We provide anonymized worker-ids.} 

For each dataset, we randomly select 250 sentence pairs as the test set and use the rest as the training set. To ensure accurate labeling, we perform an additional round of annotation on the test sets. We ask four crowdworkers to label each pair using the same instructions that we use for data labeling, giving us a total of 5 annotations per example. We assign the majority vote as the gold label. 

Table~\ref{tab:agreement} shows the agreement statistics for each protocol. \baseline\ shows a higher agreement than \sent\ and \translate,\ although it is lower than MNLI. Compared to MNLI, all of our datasets show higher number of examples with no gold label (no consensus between annotators). As we strictly follow the MNLI protocol for \baseline,\ this suggests that the different population of crowdworkers is likely responsible for these differences.\footnote{MNLI used an organized group of crowdworkers hired through Hybrid (gethybrid.io).} 

\begin{table*}[ht]
    \centering
    \small
    \begin{tabular}{llrrrrrrrrrr}
    \toprule
    & & \multicolumn{10}{c}{\textbf{Test Data}} \\
    \cmidrule{3-12}
    & \textbf{Training Data} & BN & BW & SN & SW & TW & MNLI & A1 & A2 & A3 & Avg. \\
    \midrule
    \parbox[t]{2mm}{\multirow{5}{*}{\rotatebox[origin=c]{90}{\textbf{CBoW}}}}
    & Base-News & 33.4 & 37.8 & 32.4 & 30.1 & 35.8 & 35.6 & 32.8 & 32.8 & \textbf{33.4} & 34.0 \\
    & Base-Wiki & 34.1 & 33.1 & 37.9 & 35.4 & 39.0 & 35.6 & 33.1 & 31.6 & 33.2 & 34.8 \\
    & Sim-News & 35.4 & 35.9 & 32.0 & 32.3 & 37.8 & 35.8 & 33.1 & 32.8 & \textbf{33.4} & 34.3 \\
    & Sim-Wiki & 32.3 & 37.2 & 52.1 & 49.1 & 44.6 & 36.6 & 33.1 & 32.4 & 32.1 & 38.8 \\
    & Translate-Wiki & 37.4 & 39.3 & 35.4 & 35.8 & 45.5 & 35.4 & 33.0 & \textbf{32.9} & 32.8 &  36.4 \\
    \midrule
    \parbox[t]{2mm}{\multirow{6}{*}{\rotatebox[origin=c]{90}{\textbf{RoBERTa}}}}
    & MNLI-3k & 79.0 & 61.3 & \textbf{76.7} & 57.5 & 58.1 & \textbf{83.9} & 33.4 & 27.0 & 28.7 & 56.2 \\
    \cmidrule{2-12}
    & Base-News & \textbf{79.4} & \textbf{76.1} & 57.5 & 61.6 & 58.1 & 83.1 & \textbf{35.8} & 29.5 & 28.0 & \textbf{56.6} \\
    & Base-Wiki & 77.0 & 74.2 & 58.5 & 62.0 & 61.3 & 54.0 & 30.9 & 31.8 & 33.1 & 53.6 \\
    & Sim-News & 53.3 & 56.0 & 65.8 & 59.8 & 66.2 & 79.5 & \textbf{35.8} & 30.2 & 28.2 & 52.8 \\
    & Sim-Wiki & 62.0 & 62.8 & 64.8 & \textbf{64.9} & \textbf{69.1} & 64.7 & 32.2 & 32.0 & 31.5 & 53.8 \\
    & Translate-Wiki & 48.5 & 54.9 & 60.7 & 58.1 & 67.1 & 50.9 & 32.5 & 32.7 & 33.2 & 48.7 \\
    \midrule
    \multicolumn{2}{l}{\textit{Average per test set}} & 52.0 & 51.7 & 52.2 & 49.7 & 53.0 & 54.1 & 33.2 & 31.4 & 31.6 & 45.4 \\
    \bottomrule
    \end{tabular}
    \caption{Model performance on individual test sets, as a median over 10 random restarts. \textbf{BN}: Base-News, \textbf{BW}: Base-Wiki, SN: \textbf{Sim-News}, \textbf{SW}: Sim-Wiki, \textbf{TW}: Translate-Wiki. The last row shows the average performance across models on each test set.}
    \label{tab:nli-exp}
\end{table*}

\begin{table*}[ht]
    \centering
    \small
    \begin{tabular}{llcccccccccc}
    \toprule
    & \multicolumn{10}{c}{\textbf{Test Data}} \\
    \cmidrule{2-11}
    \textbf{Training Data} & BN & BW & SN & SW & TW & MNLI & A1 & A2 & A3 & Avg. \\
    \midrule
    MNLI-3k & 46.5 & \textbf{50.4} & 33.3 & 38.4 & 36.2 & \textbf{52.8} & 33.3 & 33.1 & 33.0 & \textbf{39.7} \\
    \midrule
    Base-News & \textbf{47.8} & 46.6 & 33.8 & 33.6 & \textbf{37.4} & 51.5 & \textbf{32.5} & \textbf{33.3} & 33.1 & 38.8 \\
    Base-Wiki & 33.2 & 32.1 & \textbf{44.3} & \textbf{45.0} & 29.3 & 32.8 & 33.3 & \textbf{33.3} & 33.0 & 35.1 \\
    Sim-News & 33.2 & 35.5 & 38.8 & 38.9 & 29.3 & 32.8 & 33.3 & \textbf{33.3} & \textbf{33.5} & 34.3 \\
    Sim-Wiki & 33.2 & 30.8 & \textbf{44.3} & 44.6 & 28.8 & 32.8 & 33.3 & \textbf{33.3} & 33.0 & 34.9 \\
    Translate-Wiki & 31.4 & 34.6 & 34.3 & 34.5 & 32.4 & 33.6 & 33.3 & \textbf{33.3} & \textbf{33.5} & 33.4 \\
    \midrule
    \textit{Average per test set} & 37.5 & 38.3 & 38.1 & 39.2 & 32.2 & 39.4 & 33.2 & 33.3 & 33.2 & 36.0 \\
    \bottomrule
    \end{tabular}
    \caption{RoBERTa performance on individual test sets for \textit{hypothesis-only} models.}
    \label{tab:nli-ho-exp}
\end{table*}

\subsection{Dataset Statistics}
\label{sec:data-analysis}

Table~\ref{tab:data-stat} shows the statistics of our collected data. As anticipated, datasets collected using \sent\ and \translate\ have slightly unbalanced distributions compared to \baseline.\ In particular, for \sent,\ we observe that the entailment class has the lowest distribution in the training and test data.

One clear difference between \textsc{Base} and our new protocols is the hypothesis length. \textsc{Sim} and \textsc{Translate} tend to create longer hypothesis than \textsc{Base}. We suspect that this is an artifact of the sentence-similarity method, which prefers \textit{identical} sentences (both syntax and semantics) over semantically \textit{similar} sentences. Across domains, we observe that sentences from news texts are longer than Wikipedia.

Recent work by \citet{mccoy-etal-2019-right} shows that popular NLI models might learn a simple lexical overlap heuristic for predicting entailment labels. While this heuristic is natural for entailment, it can affect the model's generalization especially when it is strongly reflected in the data. We calculate word type overlap by using the intersection of premise and hypothesis word types, divided by the union of the two sets. The last three columns in Table~\ref{tab:data-stat} reports word type overlap in each dataset for each entailment label. We find that word type overlap is a \textit{much} stronger predictor of the label in our new protocols than in \baseline. This could be a significant driver of our results and might hurt the generalization performance of models trained using our new protocols' data.

\begin{table*}[t]
    \centering \small
    \begin{tabular}{llcccccc}
    \toprule
    \multicolumn{2}{l}{Intermediate} & COPA & MultiRC & RTE & WiC & WSC & \multirow{2}{*}{Avg.} \\
    \multicolumn{2}{l}{training data} & acc. & F1$_\alpha$ & acc. & acc. & acc. \\
    \midrule
    \parbox[t]{2mm}{\multirow{7}{*}{\rotatebox[origin=c]{90}{\textbf{BERT}}}}
    & None & 70.0 & 70.9 & 73.3 & 72.7 & 62.5 & 69.9 \\
    \cmidrule{2-8}
    & MNLI-3k & +0.0 & -0.1 & \textbf{+4.0} & -0.8 & -2.9 & +0.0 \\
    \cmidrule{2-8}
    & Base-News & \textbf{+1.0} & -0.5 & \textbf{+4.3} & -1.7 & \textbf{+1.0} & \textbf{+0.8} \\
    & Base-Wiki & \textbf{+2.0} & \textbf{+0.3} & \textbf{+3.2} & -1.2 & -1.0 & \textbf{+0.7} \\
    & Sim-News & \textbf{+3.0} & -0.3 & \textbf{+2.2} & -2.3 & +0.0 & \textbf{+0.5} \\
    & Sim-Wiki & \textbf{+7.0} & -0.2 & \textbf{+4.0} & -2.6 & -3.8 & \textbf{+0.9} \\
    & Translate-Wiki & \textbf{+4.0} & \textbf{+0.1} & \textbf{+2.5} & -3.7 & 0.0 & \textbf{+0.6} \\
    \midrule
    \midrule
    \parbox[t]{2mm}{\multirow{7}{*}{\rotatebox[origin=c]{90}{\textbf{RoBERTa}}}}
    & None & 88.0 & 77.0 & 85.2 & 71.9 & 67.3 & 77.9 \\
    \cmidrule{2-8}
    & MNLI-3k & -4.0 & -0.1 & +0.7 & +0.2 & -3.8 & -1.5 \\
    \cmidrule{2-8}
    & Base-News & \bf{+1.0} & \bf{+0.4} & \textbf{+1.1} & \textbf{+0.7} & -1.9 & \textbf{+0.3} \\
    & Base-Wiki & -2.0 & -1.2 & \textbf{+1.1} & +0.5 & -1.0 & -0.5 \\
    & Sim-News & -6.0 & -3.6 & -6.1 & -0.1 & -3.8 & -3.9 \\
    & Sim-Wiki & -5.0 & -1.9 & -2.2 & -1.2 & -16.3 & -5.3 \\
    & Translate-Wiki & -5.0 & -2.7 & -2.5 & -1.8 & -6.7 & -3.7 \\
    \bottomrule
    \end{tabular}
    \caption{Results on using each collected dataset as intermediate training data on five SuperGLUE tasks. We report the median performance over 3 random restarts on the intermediate NLI models. \textit{None} denotes experiments without intermediate-task training, i.e., direct fine-tuning on target tasks. The last column shows the average score across the five tasks. We report the difference with respect to \textit{None} using BERT and RoBERTa.}
    \label{tab:transfer-exp}
\end{table*}

\subsection{Annotation Cost}

We use the FairWork platform to set payment for each of our HITs \citep{whiting2019fair}. FairWork surveys workers to estimate the time that each HIT takes and adjusts pay to a target of US \$15/hr. Based on its estimation, we pay $\$0.4$ and $\$0.3$ for each written hypothesis of Base-News and Base-Wiki, respectively. For Sim-News, Sim-Wiki, and Translate-Wiki, we pay $\$0.175$, $\$0.15$, $\$0.15$ for each labeled sentence pair, respectively. In total, we spend $\$1791$ for each dataset collected from Gigaword and $\$1445$ for each dataset collected from Wikipedia.

\section{Experiments}

We aim to test whether our alternative protocols can produce high-quality data that yield models that generalize well within NLI and in transfer learning. For the NLI evaluation, we evaluate each model on nine test sets: (i) the five new individual test sets, each containing \mytilde250 examples; (ii) the MNLI \textit{development} set; and (iii) the three \textit{development} sets of Adversarial NLI \citep[ANLI;][]{ANLI}, collected from three rounds of annotation (A1, A2, A3). ANLI is collected using an iterative adversarial approach that follows MNLI but encourages crowdworkers to write sentences that are difficult for a trained NLI model. 

We experiment with two sentence encoders: a CBoW baseline initialized with fastText embeddings \citep{bojanowski-etal-2017-enriching}, and a more powerful RoBERTa$_{\rm{Large}}$ \citep{liu2019roberta} model, fine-tuned on individual training sets. We perform a hyperparameter sweep, varying the learning rate $\in \{1e-3, 1e-4, 1e-5\}$ and the dropout rate  $\in \{0.1, 0.2\}$. We use batch size of 16 and 4 for CBoW and RoBERTA, respectively. We train each model using the best hyperparameters for 10 epochs, with 10 random restarts. In initial experiments, we find that this setup yields sTable~performance given our relatively small datasets, especially when using RoBERTa.\footnote{This is consistent with the recent findings of \citet{zhang2020revisiting} and \citet{ mosbach2020stability} regarding fine-tuning BERT-style models on small data.}

For transfer learning, we test whether each dataset can improve downstream task performance when it is used as intermediate-task data \citep{phang2018sentence,pruksachatkun2020intermediatee}. As our collected datasets are fairly small ($<$ 10K examples), we use five data-poor downstream target tasks in the SuperGLUE benchmark \citep{wang2019superglue}: \textbf{COPA} \citep{copa}; \textbf{WSC} \citep{wsc}; \textbf{RTE} \citep[][et seq]{rte}, \textbf{WiC} \citep{wic}; and \textbf{MultiRC} \citep{multirc}. We experiment with the BERT$_{\rm{Large}}$ \citep{devlin-etal-2019-bert} and RoBERTa$_{\rm{Large}}$ models.
We follow \citet{pruksachatkun2020intermediatee} for training hyperparameters. We use the Adam optimizer \citep{KingmaB14}.

We run experiments using the \texttt{jiant} toolkit \citep{wang2019jiant}, which is the recommended baseline package for SuperGLUE, and is based on Pytorch \citep{NEURIPS2019_9015}, HuggingFace Transformers \citep{Wolf2019HuggingFacesTS}, and AllenNLP \citep{Gardner2017AllenNLP}.

\subsection{NLI Experiments}

Table~\ref{tab:nli-exp} reports the model performance on individual test sets. We include a baseline training data, a 3k randomly sampled training examples from MNLI (MNLI-3k). We observe that all the CBoW baselines obtain near chance performance. Using RoBERTa, the top performing models are all trained on datasets collected using \baseline:\ Base-News and MNLI-3k. We find that models trained using Translate-Wiki obtain the worst performance. On average across all training sets, ANLI development sets seem to be the hardest, while MNLI seems to be the easiest.

Unsurprisingly, we do not find a single training set which yields the best model across all test sets. We observe that models trained on Base-News perform the best for Base-News and Base-Wiki test sets. Similarly, Sim-Wiki performs the best on both Sim-Wiki and Sim-News test sets. We find that all models do poorly on all ANLI development sets.

Overall, we find that Base-News outperforms all other datasets. However, it is also better than \sent\ and \translate\ which suggests that our new protocols failed. The lower accuracy for \sent\ and \translate\ on their respective test sets also suggests that they produce datasets with noisier labels.

\subsection{Hypothesis-Only Results}

Next, we experiment with a hypothesis-only model \citep{poliak-etal-2018-hypothesis} to investigate spurious statistical patterns in the hypotheses which might signal the actual labels to the model. Table~\ref{tab:nli-ho-exp} reports the results for all five datasets and MNLI. On the five new test sets, we observe that MNLI and Base-News are the most solvable by the hypothesis-only models, though their numbers are still much lower than with SNLI with accuracy 69.17. 

On average across all test sets, none of the training sets obtain much higher performance than chance. All models achieve chance performance on ANLI. However, all of our training sets are fairly small, and these numbers might not be very informative. This also explains why these numbers are relatively lower than other NLI datasets \citep{poliak-etal-2018-hypothesis}. Across all training sets, we again see that the MNLI test set is the most solvable by the hypothesis-only models.

Our new protocols show lower performance than the \baseline,\ but that may just be because they are of lower overall quality and not because they are less solvable by the hypothesis-only models. We verify this by looking at their transfer learning performance in the following section.

\subsection{Transfer Learning}

Table~\ref{tab:transfer-exp} shows our results when using each collected data as intermediate-training data on the five target tasks. We report the median performance of three random restarts on the validation sets. Using BERT, we observe that all our new datasets yield models with better performance than plain BERT or MNLI-3k as intermediate-training data. We see less positive transfer when we use RoBERTa.

If we look at individual target task performance, both Base-News and Base-Wiki data give consistent positive transfer for RTE, a natural language inference task. We also see some positive transfer for COPA, however since its validation set is very small (100 examples), we can not conclude anything with confidence. 

Overall, our \baseline\ shows better transfer learning performance compared to MNLI, suggesting that our setup is sound. However, we also see that our new protocols perform worse than \baseline,\ showing that they produce less useful training data than the strong baseline of crowdworker writing.

\section{Dataset Analysis}

\subsection{Annotation Artifacts}

Following \citet{gururangan-etal-2018-annotation}, we compute the PMI between each hypothesis word and label in the training set to examine whether certain words have high associations with its inference label. For a fair comparison, we only use \mytilde3k training examples from each dataset, and sub-sample data collected using \textsc{Sim} and \textsc{Translate}. 

Table~\ref{tab:pmi-analysis} shows the top three most associated words for each label, sorted by their PMI scores. We find that \textsc{Base} has similar associations to MNLI, especially for the neutral and contradiction labels where we found many negations and adverbs. We observe that both \textsc{Sim} and \textsc{Translate} are less susceptible to this artifact. However, this might be a side-effect of high word overlap in the data, which prefers similar words in the premise and hypothesis. This is also a well-known artifact for NLI data \citep{mccoy-etal-2019-right}.

\begin{table}[t]
    \centering
    \small
    \setlength\tabcolsep{1.3mm}
    \begin{tabular}{llrlrlr}
    \toprule
     & \multicolumn{2}{c}{\textbf{Entailment}} &
     \multicolumn{2}{c}{\textbf{Contradiction}} &  \multicolumn{2}{c}{\textbf{Neutral}} \\ 
    \midrule
    \parbox[t]{2mm}{\multirow{3}{*}{\rotatebox[origin=c]{90}{M-3k}}}
    & looked & 0.44 & no & 1.03 & also & 0.75 \\
    & capital & 0.43 & never & 0.95 & because & 0.71 \\
    & population & 0.43 & any & 0.88 & better & 0.63 \\
    \midrule
    \parbox[t]{2mm}{\multirow{3}{*}{\rotatebox[origin=c]{90}{B-News}}}
    & according & 0.58 & never & 1.07 & also & 0.62 \\
    & position & 0.45 & no & 1.02 & many & 0.52 \\
    & set & 0.42 & any & 0.90 & most & 0.52 \\
    \midrule
    \parbox[t]{2mm}{\multirow{3}{*}{\rotatebox[origin=c]{90}{B-Wiki}}}
    & both & 0.45 & never & 1.18 & most & 0.78 \\
    & named & 0.38 & not & 1.01 & well & 0.64 \\
    & early & 0.35 & any & 0.96 & many & 0.56 \\
    \midrule
    \parbox[t]{2mm}{\multirow{3}{*}{\rotatebox[origin=c]{90}{S-Giga}}}
    & summit & 0.53 & points & 0.66 & very & 0.54 \\
    & roads & 0.51 & we & 0.65 & research & 0.48 \\
    & weighted & 0.46 & -- & 0.59 & weeks & 0.48 \\
    \midrule
    \parbox[t]{2mm}{\multirow{3}{*}{\rotatebox[origin=c]{90}{S-Wiki}}}
    & division & 0.56 & census & 0.88 & through & 0.57 \\
    & team & 0.48 & population & 0.86 & such & 0.54 \\
    & candidate & 0.47 & 2010 & 0.82 & number & 0.49 \\
    \midrule
    \parbox[t]{2mm}{\multirow{3}{*}{\rotatebox[origin=c]{90}{T-Wiki}}}
    & ; & 0.68 & brought & 0.45 & each & 0.57 \\
    & album & 0.58 & maintain & 0.40 & \{ & 0.56 \\
    & f & 0.55 & will & 0.39 & \} & 0.56 \\
    \bottomrule
    \end{tabular}
    \caption{Top three words most associated with each label by PMI. \textbf{M}: MNLI, \textbf{B}: Base, \textbf{S}: Sim, \textbf{T}: Translate.}
    \label{tab:pmi-analysis}
\end{table}

\subsection{Qualitative Analysis}

Our new protocols use a vector-distance based measurement to find similar sentences, and we find that many of the sentence pairs share similar syntactic structure in their premise and hypothesis, even when both describe different events or entities. We also find that hypothesis in several Sim-News examples differs by only a few words with its premise. For Translate-Wiki, we observe some effects of \textit{translation divergence}, where the translation of the sentence changes semantically because of cross-linguistic distinctions between languages. We provide some examples of these observations in Table~\ref{tab:qualitative-analysis}.

\begin{table*}[ht]
    \centering
    \small
    \newcolumntype{L}{>{\arraybackslash}m{.3\textwidth}}
    \newcolumntype{K}{>{\arraybackslash}m{.1\textwidth}}
    \begin{tabular}{KlLLc}
    \toprule
    \textbf{Type} & \textbf{Dataset} & \textbf{Premise} & \textbf{Hypothesis} & \textbf{Label} \\
    \midrule
    Syntactic structure 
        & Sim-News
        & \textbf{For many people}, choosing wallpaper is one of decorating's more stressful experiences, fraught with anxiety over color, pattern and cost.
        & \textbf{For many people}, anxiety about decorating stems from not understanding the language of furniture, fabrics and decorative styles.
        & E \\
        \addlinespace[.15cm]
        & Sim-Wiki
        & \textbf{Its flowers are} pale yellow \textbf{to} white \textbf{and} spherical.
        & \textbf{Its flowers are} funnel-shaped \textbf{and} pink \textbf{to} white. 
        & C \\
        \addlinespace[.15cm]
        & Translate-Wiki
        & But now, in \textbf{the early 1990s}, the Jakarta-Begor railway had turned into a double rail.
        & However, by \textbf{the early 1990s}, McCreery's position within the UDA became less secure.
        & N \\
    \midrule
    Lexical overlap 
     & Sim-News 
     & \textbf{GrandMet owns Burger King, the world's second-biggest hamburger chain, as well as US} frozen foods manufacturer \textbf{Pillsbury, which produces the luxury ice-cream Haagen-Daazs.}
     & \textbf{GrandMet owns Burger King, the world's second-biggest hamburger chain, as well as US} food group \textbf{Pillsbury, which produces the luxury ice-cream Haagen-Daazs.}
     & E \\
    \midrule
    Translation divergence & Translate-Wiki 
    & Marcus Claudius then abducted \textbf{her} while \textbf{she} was on \textbf{her} way to school.
    & Marcus Claudius then kidnapped \textbf{him} while \textbf{he} was on \textbf{his} way to school.
    & N \\
    \bottomrule
    \end{tabular}
    \caption{Dataset observations from our new protocols.}
    \label{tab:qualitative-analysis}
\end{table*}

\section{Related Work}

There is a large body of work on constructing data for natural language inference. The first test suite for entailment problems, FraCas \citep{Consortium96usingthe}, is a very small set created manually by experts to isolate phenomena of interest. The RTE challenge corpora \citep[][{et seq}]{rte} were built by asking human annotators to judge whether a text entails a hypothesis. The SICK dataset \citep{marelli-etal-2014-sick} is constructed by mining existing paraphrase sentence pairs from image and video captions, which annotators then label. 

Some recent works also use automatic methods for generating sentence pairs for entailment data. \citet{zhang-etal-2017-ordinal} propose a framework to generate hypotheses based on context from general world knowledge or neural sequence-to-sequence methods. The DNC corpus \citep{poliak-etal-2018-collecting} is an NLI dataset with ordinal judgments constructed by recasting several NLP datasets to NLI examples and labeling them using custom automatic procedures. QA-NLI \citep{QA-NLI} is an NLI dataset derived from existing QA datasets. Similar to ours, both DNC and QA-NLI use automatic methods to generate sentence pairs. However, neither of them explicitly evaluates whether machine-generated pairs are better than human-generated pairs.

\citet{bowman2020collecting} propose four potential modifications to the SNLI/MNLI protocol, all still involving crowdworker writing, and show that none yields improvements in the resulting data. SWAG \citep{zellers-etal-2018-swag} and HellaSwag \citep{zellers-etal-2019-hellaswag} construct sentence pairs from specific data sources and use language models to generate challenging negative examples. 

On the topic of cost-effective crowdsourcing, \citet{gao-etal-2015-cost} develop a method to reduce redundant translations when collecting human translated data. When the annotation budget is fixed, \citet{khetan2018learning} suggest that it is better to label collect single label per training example as many as possible, rather than collecting less training examples with multiple labels.

\section{Conclusion}

In this paper, we introduce two data collection protocols which use fully-automatic pipelines to collect hypotheses, replacing crowdworker writing in the MNLI baseline protocol. We find that switching to a writing-free process with the same source data and annotator pool yields poor-quality data. Our main experiments show strong negative results both in NLI generalization and transfer learning, and mixed results on annotation artifacts, suggesting that MNLI-style crowdworker writing examples are broadly better than automatically paired ones. This finding dovetails with that of \citet{bowman2020collecting}, who find that they are unable to improve upon a base MNLI-style prompt when introducing aids meant to improve annotator speed or creativity. Future work along this line might focus on crowdsourcing strategies (beyond the basic HIT design) which encourage crowdworkers to produce high-quality data with reduced artifacts.

While our fully-automatic methods to construct sentence pairs yield negative results, we have not exhausted all possible automatic techniques for collecting similar sentences. However, given that we use state-of-the-art tools including FAISS, RoBERTa, and OPUS, and refine our methods with several rounds of piloting and tuning, we are skeptical that there is low-hanging fruit in the two directions we explored. A more radically different direction might involve generating pairs from scratch, using a large language model like GPT-3 \citep{brown2020language}. However, this would still require training data from crowdworker-written dataset, and might add a major source of potentially difficult-to-diagnose bias.

Finally, despite its known issues, we find that MNLI-style data is still the most effective for both NLI evaluation and transfer learning, and future efforts to create similar data should work from that starting point.

\section*{Acknowledgments}
This project has benefited from financial support to SB by Eric and Wendy Schmidt (made by recommendation of the Schmidt Futures program), by Samsung Research (under the project \textit{Improving Deep Learning using Latent Structure}), by Intuit, Inc., and in-kind support by the NYU High-Performance Computing Center and by NVIDIA Corporation (with the donation of a Titan V GPU). This material is based upon work supported by the National Science Foundation under Grant No. 1922658. Any opinions, findings, and conclusions or recommendations expressed in this material are those of the author(s) and do not necessarily reflect the views of the National Science Foundation.

\bibliography{anthology,aacl-ijcnlp2020}

\begin{thebibliography}{48}
\expandafter\ifx\csname natexlab\endcsname\relax\def\natexlab#1{#1}\fi

\bibitem[{Akiba et~al.(2019)Akiba, Sano, Yanase, Ohta, and
  Koyama}]{optuna_2019}
Takuya Akiba, Shotaro Sano, Toshihiko Yanase, Takeru Ohta, and Masanori Koyama.
  2019.
\newblock {Optuna: A Next-generation Hyperparameter Optimization Framework}.
\newblock In \emph{Proceedings of the 25rd {ACM} {SIGKDD} International
  Conference on Knowledge Discovery and Data Mining}.

\bibitem[{Artetxe and Schwenk(2019)}]{artetxe-schwenk-2019-margin}
Mikel Artetxe and Holger Schwenk. 2019.
\newblock \href {https://doi.org/10.18653/v1/P19-1309} {Margin-based parallel
  corpus mining with multilingual sentence embeddings}.
\newblock In \emph{Proceedings of the 57th Annual Meeting of the Association
  for Computational Linguistics}, pages 3197--3203, Florence, Italy.
  Association for Computational Linguistics.

\bibitem[{Bojanowski et~al.(2017)Bojanowski, Grave, Joulin, and
  Mikolov}]{bojanowski-etal-2017-enriching}
Piotr Bojanowski, Edouard Grave, Armand Joulin, and Tomas Mikolov. 2017.
\newblock \href {https://doi.org/10.1162/tacl_a_00051} {Enriching word vectors
  with subword information}.
\newblock \emph{Transactions of the Association for Computational Linguistics},
  5:135--146.

\bibitem[{Bowman et~al.(2015)Bowman, Angeli, Potts, and
  Manning}]{bowman-etal-2015-large}
Samuel~R. Bowman, Gabor Angeli, Christopher Potts, and Christopher~D. Manning.
  2015.
\newblock \href {https://doi.org/10.18653/v1/D15-1075} {A large annotated
  corpus for learning natural language inference}.
\newblock In \emph{Proceedings of the 2015 Conference on Empirical Methods in
  Natural Language Processing}, pages 632--642, Lisbon, Portugal. Association
  for Computational Linguistics.

\bibitem[{Bowman et~al.(2020)Bowman, Palomaki, Soares, and
  Pitler}]{bowman2020collecting}
Samuel~R. Bowman, Jennimaria Palomaki, Livio~Baldini Soares, and Emily Pitler.
  2020.
\newblock \href {http://arxiv.org/abs/2004.11997} {{Collecting Entailment Data
  for Pretraining: New Protocols and Negative Results}}.

\bibitem[{Brown et~al.(2020)Brown, Mann, Ryder, Subbiah, Kaplan, Dhariwal,
  Neelakantan, Shyam, Sastry, Askell, Agarwal, Herbert-Voss, Krueger, Henighan,
  Child, Ramesh, Ziegler, Wu, Winter, Hesse, Chen, Sigler, Litwin, Gray, Chess,
  Clark, Berner, McCandlish, Radford, Sutskever, and
  Amodei}]{brown2020language}
Tom~B. Brown, Benjamin Mann, Nick Ryder, Melanie Subbiah, Jared Kaplan,
  Prafulla Dhariwal, Arvind Neelakantan, Pranav Shyam, Girish Sastry, Amanda
  Askell, Sandhini Agarwal, Ariel Herbert-Voss, Gretchen Krueger, Tom Henighan,
  Rewon Child, Aditya Ramesh, Daniel~M. Ziegler, Jeffrey Wu, Clemens Winter,
  Christopher Hesse, Mark Chen, Eric Sigler, Mateusz Litwin, Scott Gray,
  Benjamin Chess, Jack Clark, Christopher Berner, Sam McCandlish, Alec Radford,
  Ilya Sutskever, and Dario Amodei. 2020.
\newblock \href {http://arxiv.org/abs/2005.14165} {{Language Models are
  Few-Shot Learners}}.

\bibitem[{Conneau et~al.(2018)Conneau, Rinott, Lample, Williams, Bowman,
  Schwenk, and Stoyanov}]{conneau-etal-2018-xnli}
Alexis Conneau, Ruty Rinott, Guillaume Lample, Adina Williams, Samuel Bowman,
  Holger Schwenk, and Veselin Stoyanov. 2018.
\newblock \href {https://doi.org/10.18653/v1/D18-1269} {{XNLI}: Evaluating
  cross-lingual sentence representations}.
\newblock In \emph{Proceedings of the 2018 Conference on Empirical Methods in
  Natural Language Processing}, pages 2475--2485, Brussels, Belgium.
  Association for Computational Linguistics.

\bibitem[{Consortium et~al.(1996)Consortium, Cooper, Crouch, Eijck, Fox,
  Genabith, Jaspars, Kamp, Milward, Pinkal, Poesio, Pulman, Briscoe, Maier, and
  Konrad}]{Consortium96usingthe}
The~Fracas Consortium, Robin Cooper, Dick Crouch, Jan~Van Eijck, Chris Fox,
  Josef~Van Genabith, Jan Jaspars, Hans Kamp, David Milward, Manfred Pinkal,
  Massimo Poesio, Steve Pulman, Ted Briscoe, Holger Maier, and Karsten Konrad.
  1996.
\newblock {Using the Framework}.

\bibitem[{Dagan et~al.(2005)Dagan, Glickman, and Magnini}]{rte}
Ido Dagan, Oren Glickman, and Bernardo Magnini. 2005.
\newblock The pascal recognising textual entailment challenge.
\newblock In \emph{Machine Learning Challenges Workshop}, pages 177--190.
  Springer.

\bibitem[{Demszky et~al.(2018)Demszky, Guu, and Liang}]{QA-NLI}
Dorottya Demszky, Kelvin Guu, and Percy Liang. 2018.
\newblock \href {http://arxiv.org/abs/1809.02922} {{Transforming Question
  Answering Datasets Into Natural Language Inference Datasets}}.

\bibitem[{Devlin et~al.(2019)Devlin, Chang, Lee, and
  Toutanova}]{devlin-etal-2019-bert}
Jacob Devlin, Ming-Wei Chang, Kenton Lee, and Kristina Toutanova. 2019.
\newblock \href {https://doi.org/10.18653/v1/N19-1423} {{BERT}: Pre-training of
  deep bidirectional transformers for language understanding}.
\newblock In \emph{Proceedings of the 2019 Conference of the North {A}merican
  Chapter of the Association for Computational Linguistics: Human Language
  Technologies, Volume 1 (Long and Short Papers)}, pages 4171--4186,
  Minneapolis, Minnesota. Association for Computational Linguistics.

\bibitem[{Gao et~al.(2015)Gao, Xu, and Callison-Burch}]{gao-etal-2015-cost}
Mingkun Gao, Wei Xu, and Chris Callison-Burch. 2015.
\newblock \href {https://doi.org/10.3115/v1/N15-1072} {Cost optimization in
  crowdsourcing translation: Low cost translations made even cheaper}.
\newblock In \emph{Proceedings of the 2015 Conference of the North {A}merican
  Chapter of the Association for Computational Linguistics: Human Language
  Technologies}, pages 705--713, Denver, Colorado. Association for
  Computational Linguistics.

\bibitem[{Gardner et~al.(2017)Gardner, Grus, Neumann, Tafjord, Dasigi, Liu,
  Peters, Schmitz, and Zettlemoyer}]{Gardner2017AllenNLP}
Matt Gardner, Joel Grus, Mark Neumann, Oyvind Tafjord, Pradeep Dasigi,
  Nelson~F. Liu, Matthew Peters, Michael Schmitz, and Luke~S. Zettlemoyer.
  2017.
\newblock \href {https://arxiv.org/abs/1803.07640} {{AllenNLP: A Deep Semantic
  Natural Language Processing Platform}}.
\newblock Unpublished manuscript available on arXiv.

\bibitem[{Gururangan et~al.(2018)Gururangan, Swayamdipta, Levy, Schwartz,
  Bowman, and Smith}]{gururangan-etal-2018-annotation}
Suchin Gururangan, Swabha Swayamdipta, Omer Levy, Roy Schwartz, Samuel Bowman,
  and Noah~A. Smith. 2018.
\newblock \href {https://doi.org/10.18653/v1/N18-2017} {Annotation artifacts in
  natural language inference data}.
\newblock In \emph{Proceedings of the 2018 Conference of the North {A}merican
  Chapter of the Association for Computational Linguistics: Human Language
  Technologies, Volume 2 (Short Papers)}, pages 107--112, New Orleans,
  Louisiana. Association for Computational Linguistics.

\bibitem[{{Johnson} et~al.(2019){Johnson}, {Douze}, and {Jégou}}]{FAISS}
J.~{Johnson}, M.~{Douze}, and H.~{Jégou}. 2019.
\newblock \href {https://doi.org/10.1109/TBDATA.2019.2921572} {{Billion-scale
  similarity search with GPUs}}.
\newblock \emph{IEEE Transactions on Big Data}.

\bibitem[{Khashabi et~al.(2018)Khashabi, Chaturvedi, Roth, Upadhyay, and
  Roth}]{multirc}
Daniel Khashabi, Snigdha Chaturvedi, Michael Roth, Shyam Upadhyay, and Dan
  Roth. 2018.
\newblock \href {https://doi.org/10.18653/v1/N18-1023} {{Looking Beyond the
  Surface: A Challenge Set for Reading Comprehension over Multiple Sentences}}.
\newblock In \emph{Proceedings of the 2018 Conference of the North {A}merican
  Chapter of the Association for Computational Linguistics: Human Language
  Technologies, Volume 1 (Long Papers)}, pages 252--262, New Orleans,
  Louisiana. Association for Computational Linguistics.

\bibitem[{Khetan et~al.(2018)Khetan, Lipton, and
  Anandkumar}]{khetan2018learning}
Ashish Khetan, Zachary~C. Lipton, and Anima Anandkumar. 2018.
\newblock \href {https://openreview.net/forum?id=H1sUHgb0Z} {{Learning From
  Noisy Singly-labeled Data}}.
\newblock In \emph{International Conference on Learning Representations}.

\bibitem[{Kingma and Ba(2015)}]{KingmaB14}
Diederik~P. Kingma and Jimmy Ba. 2015.
\newblock \href {http://arxiv.org/abs/1412.6980} {{Adam: {A} Method for
  Stochastic Optimization}}.
\newblock In \emph{3rd International Conference on Learning Representations,
  {ICLR} 2015, San Diego, CA, USA, May 7-9, 2015, Conference Track
  Proceedings}.

\bibitem[{Levesque et~al.(2012)Levesque, Davis, and Morgenstern}]{wsc}
Hector~J. Levesque, Ernest Davis, and Leora Morgenstern. 2012.
\newblock \href {http://dl.acm.org/citation.cfm?id=3031843.3031909} {{The
  Winograd Schema Challenge}}.
\newblock In \emph{Proceedings of the Thirteenth International Conference on
  Principles of Knowledge Representation and Reasoning}, KR'12, pages 552--561.
  AAAI Press.

\bibitem[{Liu et~al.(2019)Liu, Ott, Goyal, Du, Joshi, Chen, Levy, Lewis,
  Zettlemoyer, and Stoyanov}]{liu2019roberta}
Yinhan Liu, Myle Ott, Naman Goyal, Jingfei Du, Mandar Joshi, Danqi Chen, Omer
  Levy, Mike Lewis, Luke Zettlemoyer, and Veselin Stoyanov. 2019.
\newblock \href {http://arxiv.org/abs/1907.11692} {{RoBERTa: A Robustly
  Optimized BERT Pretraining Approach}}.

\bibitem[{Marelli et~al.(2014)Marelli, Menini, Baroni, Bentivogli, Bernardi,
  and Zamparelli}]{marelli-etal-2014-sick}
Marco Marelli, Stefano Menini, Marco Baroni, Luisa Bentivogli, Raffaella
  Bernardi, and Roberto Zamparelli. 2014.
\newblock \href
  {http://www.lrec-conf.org/proceedings/lrec2014/pdf/363_Paper.pdf} {A {SICK}
  cure for the evaluation of compositional distributional semantic models}.
\newblock In \emph{Proceedings of the Ninth International Conference on
  Language Resources and Evaluation ({LREC}-2014)}, pages 216--223, Reykjavik,
  Iceland. European Languages Resources Association (ELRA).

\bibitem[{McCoy et~al.(2019)McCoy, Pavlick, and Linzen}]{mccoy-etal-2019-right}
Tom McCoy, Ellie Pavlick, and Tal Linzen. 2019.
\newblock \href {https://doi.org/10.18653/v1/P19-1334} {Right for the wrong
  reasons: Diagnosing syntactic heuristics in natural language inference}.
\newblock In \emph{Proceedings of the 57th Annual Meeting of the Association
  for Computational Linguistics}, pages 3428--3448, Florence, Italy.
  Association for Computational Linguistics.

\bibitem[{Mosbach et~al.(2020)Mosbach, Andriushchenko, and
  Klakow}]{mosbach2020stability}
Marius Mosbach, Maksym Andriushchenko, and Dietrich Klakow. 2020.
\newblock \href {http://arxiv.org/abs/2006.04884} {{On the Stability of
  Fine-tuning BERT: Misconceptions, Explanations, and Strong Baselines}}.

\bibitem[{Nie et~al.(2020)Nie, Williams, Dinan, Bansal, Weston, and
  Kiela}]{ANLI}
Yixin Nie, Adina Williams, Emily Dinan, Mohit Bansal, Jason Weston, and Douwe
  Kiela. 2020.
\newblock \href {https://doi.org/10.18653/v1/2020.acl-main.441} {Adversarial
  {NLI}: A new benchmark for natural language understanding}.
\newblock In \emph{Proceedings of the 58th Annual Meeting of the Association
  for Computational Linguistics}, pages 4885--4901, Online. Association for
  Computational Linguistics.

\bibitem[{Parker et~al.(2011)Parker, Graff, Kong, Chen, and Maeda}]{Gigaword}
Robert Parker, David Graff, Junbo Kong, Ke~Chen, and Kazuaki Maeda. 2011.
\newblock {English Gigaword Fifth Edition LDC2011T07}.

\bibitem[{Paszke et~al.(2019)Paszke, Gross, Massa, Lerer, Bradbury, Chanan,
  Killeen, Lin, Gimelshein, Antiga, Desmaison, Kopf, Yang, DeVito, Raison,
  Tejani, Chilamkurthy, Steiner, Fang, Bai, and Chintala}]{NEURIPS2019_9015}
Adam Paszke, Sam Gross, Francisco Massa, Adam Lerer, James Bradbury, Gregory
  Chanan, Trevor Killeen, Zeming Lin, Natalia Gimelshein, Luca Antiga, Alban
  Desmaison, Andreas Kopf, Edward Yang, Zachary DeVito, Martin Raison, Alykhan
  Tejani, Sasank Chilamkurthy, Benoit Steiner, Lu~Fang, Junjie Bai, and Soumith
  Chintala. 2019.
\newblock \href
  {http://papers.neurips.cc/paper/9015-pytorch-an-imperative-style-high-performance-deep-learning-library.pdf}
  {{PyTorch: An Imperative Style, High-Performance Deep Learning Library}}.
\newblock In H.~Wallach, H.~Larochelle, A.~Beygelzimer, F.~d'~Alch\'{e}-Buc,
  E.~Fox, and R.~Garnett, editors, \emph{Advances in Neural Information
  Processing Systems 32}, pages 8024--8035. Curran Associates, Inc.

\bibitem[{Phang et~al.(2018)Phang, Févry, and Bowman}]{phang2018sentence}
Jason Phang, Thibault Févry, and Samuel~R. Bowman. 2018.
\newblock \href {http://arxiv.org/abs/1811.01088} {{Sentence Encoders on
  STILTs: Supplementary Training on Intermediate Labeled-data Tasks}}.

\bibitem[{Pilehvar and Camacho-Collados(2019)}]{wic}
Mohammad~Taher Pilehvar and Jose Camacho-Collados. 2019.
\newblock \href {https://doi.org/10.18653/v1/N19-1128} {{{W}i{C}: the
  Word-in-Context Dataset for Evaluating Context-Sensitive Meaning
  Representations}}.
\newblock In \emph{Proceedings of the 2019 Conference of the North {A}merican
  Chapter of the Association for Computational Linguistics: Human Language
  Technologies, Volume 1 (Long and Short Papers)}, pages 1267--1273,
  Minneapolis, Minnesota. Association for Computational Linguistics.

\bibitem[{Poliak et~al.(2018{\natexlab{a}})Poliak, Haldar, Rudinger, Hu,
  Pavlick, White, and Van~Durme}]{poliak-etal-2018-collecting}
Adam Poliak, Aparajita Haldar, Rachel Rudinger, J.~Edward Hu, Ellie Pavlick,
  Aaron~Steven White, and Benjamin Van~Durme. 2018{\natexlab{a}}.
\newblock \href {https://doi.org/10.18653/v1/D18-1007} {Collecting diverse
  natural language inference problems for sentence representation evaluation}.
\newblock In \emph{Proceedings of the 2018 Conference on Empirical Methods in
  Natural Language Processing}, pages 67--81, Brussels, Belgium. Association
  for Computational Linguistics.

\bibitem[{Poliak et~al.(2018{\natexlab{b}})Poliak, Naradowsky, Haldar,
  Rudinger, and Van~Durme}]{poliak-etal-2018-hypothesis}
Adam Poliak, Jason Naradowsky, Aparajita Haldar, Rachel Rudinger, and Benjamin
  Van~Durme. 2018{\natexlab{b}}.
\newblock \href {https://doi.org/10.18653/v1/S18-2023} {Hypothesis only
  baselines in natural language inference}.
\newblock In \emph{Proceedings of the Seventh Joint Conference on Lexical and
  Computational Semantics}, pages 180--191, New Orleans, Louisiana. Association
  for Computational Linguistics.

\bibitem[{Portelli et~al.(2020)Portelli, Zhao, Schuster, Serra, and
  Santus}]{portelli-etal-2020-distilling}
Beatrice Portelli, Jason Zhao, Tal Schuster, Giuseppe Serra, and Enrico Santus.
  2020.
\newblock \href {https://www.aclweb.org/anthology/2020.fever-1.7} {{Distilling
  the Evidence to Augment Fact Verification Models}}.
\newblock In \emph{Proceedings of the Third Workshop on Fact Extraction and
  VERification (FEVER)}, pages 47--51, Online. Association for Computational
  Linguistics.

\bibitem[{Pruksachatkun et~al.(2020)Pruksachatkun, Phang, Liu, Htut, Zhang,
  Pang, Vania, Kann, and Bowman}]{pruksachatkun2020intermediatee}
Yada Pruksachatkun, Jason Phang, Haokun Liu, Phu~Mon Htut, Xiaoyi Zhang,
  Richard~Yuanzhe Pang, Clara Vania, Katharina Kann, and Samuel~R. Bowman.
  2020.
\newblock \href {https://www.aclweb.org/anthology/2020.acl-main.467}
  {Intermediate-task transfer learning with pretrained language models: When
  and why does it work?}
\newblock In \emph{Proceedings of the 58th Annual Meeting of the Association
  for Computational Linguistics}, pages 5231--5247, Online. Association for
  Computational Linguistics.

\bibitem[{Rajpurkar et~al.(2018)Rajpurkar, Jia, and
  Liang}]{rajpurkar-etal-2018-know}
Pranav Rajpurkar, Robin Jia, and Percy Liang. 2018.
\newblock \href {https://doi.org/10.18653/v1/P18-2124} {Know what you don{'}t
  know: Unanswerable questions for {SQ}u{AD}}.
\newblock In \emph{Proceedings of the 56th Annual Meeting of the Association
  for Computational Linguistics (Volume 2: Short Papers)}, pages 784--789,
  Melbourne, Australia. Association for Computational Linguistics.

\bibitem[{Roemmele et~al.(2011)Roemmele, Bejan, and Gordon}]{copa}
Melissa Roemmele, Cosmin~Adrian Bejan, and Andrew~S Gordon. 2011.
\newblock {Choice of Plausible Alternatives: An evaluation of commonsense
  causal reasoning}.
\newblock In \emph{2011 AAAI Spring Symposium Series}.

\bibitem[{Schwenk(2018)}]{schwenk-2018-filtering}
Holger Schwenk. 2018.
\newblock \href {https://doi.org/10.18653/v1/P18-2037} {Filtering and mining
  parallel data in a joint multilingual space}.
\newblock In \emph{Proceedings of the 56th Annual Meeting of the Association
  for Computational Linguistics (Volume 2: Short Papers)}, pages 228--234,
  Melbourne, Australia. Association for Computational Linguistics.

\bibitem[{Schwenk et~al.(2019)Schwenk, Chaudhary, Sun, Gong, and
  Guzm{\'{a}}n}]{wikimatrix}
Holger Schwenk, Vishrav Chaudhary, Shuo Sun, Hongyu Gong, and Francisco
  Guzm{\'{a}}n. 2019.
\newblock \href {http://arxiv.org/abs/1907.05791} {{WikiMatrix: Mining 135M
  Parallel Sentences in 1620 Language Pairs from Wikipedia}}.
\newblock \emph{CoRR}, abs/1907.05791.

\bibitem[{Tiedemann and Thottingal(2020)}]{TiedemannThottingal:EAMT2020}
J{\"o}rg Tiedemann and Santhosh Thottingal. 2020.
\newblock {{OPUS-MT} — {B}uilding open translation services for the {W}orld}.
\newblock In \emph{Proceedings of the 22nd Annual Conferenec of the European
  Association for Machine Translation (EAMT)}, Lisbon, Portugal.

\bibitem[{Trivedi et~al.(2019)Trivedi, Kwon, Khot, Sabharwal, and
  Balasubramanian}]{trivedi-etal-2019-repurposing}
Harsh Trivedi, Heeyoung Kwon, Tushar Khot, Ashish Sabharwal, and Niranjan
  Balasubramanian. 2019.
\newblock \href {https://doi.org/10.18653/v1/N19-1302} {Repurposing entailment
  for multi-hop question answering tasks}.
\newblock In \emph{Proceedings of the 2019 Conference of the North {A}merican
  Chapter of the Association for Computational Linguistics: Human Language
  Technologies, Volume 1 (Long and Short Papers)}, pages 2948--2958,
  Minneapolis, Minnesota. Association for Computational Linguistics.

\bibitem[{Tsuchiya(2018)}]{tsuchiya-2018-performance}
Masatoshi Tsuchiya. 2018.
\newblock \href {https://www.aclweb.org/anthology/L18-1239} {Performance impact
  caused by hidden bias of training data for recognizing textual entailment}.
\newblock In \emph{Proceedings of the Eleventh International Conference on
  Language Resources and Evaluation ({LREC}-2018)}, Miyazaki, Japan. European
  Languages Resources Association (ELRA).

\bibitem[{Wang et~al.(2019{\natexlab{a}})Wang, Pruksachatkun, Nangia, Singh,
  Michael, Hill, Levy, and Bowman}]{wang2019superglue}
Alex Wang, Yada Pruksachatkun, Nikita Nangia, Amanpreet Singh, Julian Michael,
  Felix Hill, Omer Levy, and Samuel Bowman. 2019{\natexlab{a}}.
\newblock Super{GLUE}: {A} stickier benchmark for general-purpose language
  understanding systems.
\newblock In \emph{Advances in Neural Information Processing Systems}.

\bibitem[{Wang et~al.(2019{\natexlab{b}})Wang, Tenney, Pruksachatkun, Yeres,
  Phang, Liu, Htut, Yu, Hula, Xia, Pappagari, Jin, McCoy, Patel, Huang, Grave,
  Kim, F\'evry, Chen, Nangia, Mohananey, Kann, Bordia, Patry, Benton, Pavlick,
  and Bowman}]{wang2019jiant}
Alex Wang, Ian~F. Tenney, Yada Pruksachatkun, Phil Yeres, Jason Phang, Haokun
  Liu, Phu~Mon Htut, Katherin Yu, Jan Hula, Patrick Xia, Raghu Pappagari,
  Shuning Jin, R.~Thomas McCoy, Roma Patel, Yinghui Huang, Edouard Grave,
  Najoung Kim, Thibault F\'evry, Berlin Chen, Nikita Nangia, Anhad Mohananey,
  Katharina Kann, Shikha Bordia, Nicolas Patry, David Benton, Ellie Pavlick,
  and Samuel~R. Bowman. 2019{\natexlab{b}}.
\newblock {\texttt{jiant} 1.3: A software toolkit for research on
  general-purpose text understanding models}.
\newblock \url{http://jiant.info/}.

\bibitem[{Whiting et~al.(2019)Whiting, Hugh, and Bernstein}]{whiting2019fair}
Mark~E Whiting, Grant Hugh, and Michael~S Bernstein. 2019.
\newblock {Fair Work: Crowd Work Minimum Wage with One Line of Code}.
\newblock In \emph{Proceedings of the AAAI Conference on Human Computation and
  Crowdsourcing}, pages 197--206.

\bibitem[{Williams et~al.(2018)Williams, Nangia, and
  Bowman}]{williams-etal-2018-broad}
Adina Williams, Nikita Nangia, and Samuel Bowman. 2018.
\newblock \href {https://doi.org/10.18653/v1/N18-1101} {A broad-coverage
  challenge corpus for sentence understanding through inference}.
\newblock In \emph{Proceedings of the 2018 Conference of the North {A}merican
  Chapter of the Association for Computational Linguistics: Human Language
  Technologies, Volume 1 (Long Papers)}, pages 1112--1122, New Orleans,
  Louisiana. Association for Computational Linguistics.

\bibitem[{Wolf et~al.(2020)Wolf, Debut, Sanh, Chaumond, Delangue, Moi, Cistac,
  Rault, Louf, Funtowicz, Davison, Shleifer, von Platen, Ma, Jernite, Plu, Xu,
  Scao, Gugger, Drame, Lhoest, and Rush}]{Wolf2019HuggingFacesTS}
Thomas Wolf, Lysandre Debut, Victor Sanh, Julien Chaumond, Clement Delangue,
  Anthony Moi, Pierric Cistac, Tim Rault, Rémi Louf, Morgan Funtowicz, Joe
  Davison, Sam Shleifer, Patrick von Platen, Clara Ma, Yacine Jernite, Julien
  Plu, Canwen Xu, Teven~Le Scao, Sylvain Gugger, Mariama Drame, Quentin Lhoest,
  and Alexander~M. Rush. 2020.
\newblock \href {http://arxiv.org/abs/1910.03771} {Huggingface's transformers:
  State-of-the-art natural language processing}.

\bibitem[{Zellers et~al.(2018)Zellers, Bisk, Schwartz, and
  Choi}]{zellers-etal-2018-swag}
Rowan Zellers, Yonatan Bisk, Roy Schwartz, and Yejin Choi. 2018.
\newblock \href {https://doi.org/10.18653/v1/D18-1009} {{SWAG}: A large-scale
  adversarial dataset for grounded commonsense inference}.
\newblock In \emph{Proceedings of the 2018 Conference on Empirical Methods in
  Natural Language Processing}, pages 93--104, Brussels, Belgium. Association
  for Computational Linguistics.

\bibitem[{Zellers et~al.(2019)Zellers, Holtzman, Bisk, Farhadi, and
  Choi}]{zellers-etal-2019-hellaswag}
Rowan Zellers, Ari Holtzman, Yonatan Bisk, Ali Farhadi, and Yejin Choi. 2019.
\newblock \href {https://doi.org/10.18653/v1/P19-1472} {{H}ella{S}wag: Can a
  machine really finish your sentence?}
\newblock In \emph{Proceedings of the 57th Annual Meeting of the Association
  for Computational Linguistics}, pages 4791--4800, Florence, Italy.
  Association for Computational Linguistics.

\bibitem[{Zhang et~al.(2017)Zhang, Rudinger, Duh, and
  Van~Durme}]{zhang-etal-2017-ordinal}
Sheng Zhang, Rachel Rudinger, Kevin Duh, and Benjamin Van~Durme. 2017.
\newblock \href {https://doi.org/10.1162/tacl_a_00068} {Ordinal common-sense
  inference}.
\newblock \emph{Transactions of the Association for Computational Linguistics},
  5:379--395.

\bibitem[{Zhang et~al.(2020)Zhang, Wu, Katiyar, Weinberger, and
  Artzi}]{zhang2020revisiting}
Tianyi Zhang, Felix Wu, Arzoo Katiyar, Kilian~Q. Weinberger, and Yoav Artzi.
  2020.
\newblock \href {http://arxiv.org/abs/2006.05987} {{Revisiting Few-sample BERT
  Fine-tuning}}.

\end{thebibliography}
\bibliographystyle{acl_natbib}

\clearpage

\appendix

\section{Appendices}
\label{sec:appendix}

\subsection{Indexing and Retrieval}
\label{app:indexing}

\paragraph{Gigaword} The corpus contains texts from seven news sources: afp\_eng, apw\_eng, cna\_eng, ltw\_eng, nyt\_eng, wpb\_eng, and xin\_eng. We build one index for each news source with type ``\texttt{PCAR64,IVFx,Flat}", where \texttt{x} defines the number of clusters in the index. This type of index allows faster retrieval, however it requires a training stage to assign a centroid to each cluster. We refer readers to FAISS documentation for more detail explanations.\footnote{\url{https://github.com/facebookresearch/faiss}}

For each news source, we randomly sample 100 sentences from its monthly articles and use them as seed sentences to train the clusters. We then set the number of clusters \texttt{x} to $\frac{N}{100}$ (rounded to the nearest hundred), where $N$ is the number of seed sentences. Table~\ref{tab:giga-index} lists the number of seed sentences and clusters used for each news source index. 

\begin{table}[h]
    \centering
    \begin{tabular}{lrr}
    \toprule
    Source & \#seed sentences & \texttt{x} \\
    \midrule
    afp\_eng & 111,147 & 1,100 \\ 
    apw\_eng & 146,119 & 1,400 \\ 
    cna\_eng & 125,508 & 1,200 \\ 
    ltw\_eng & 90,195 & 900 \\ 
    nyt\_eng & 136,827 & 1,300 \\ 
    wpb\_eng & 9,144 & 100 \\ 
    xin\_eng & 157,760 & 1,500 \\
    \bottomrule 
    \end{tabular}
    \caption{Number of seed sentences and number of clusters for each news source index.}
    \label{tab:giga-index}
\end{table}

During retrieval, for each query, we retrieve top 1000 sentences from each index and perform reranking on the combined list, i.e., 7,000 sentence pairs, as described in Section \ref{sec:sentsim}.

\paragraph{Wikipedia} We build one index for the whole Wikipedia corpus. For seed sentences, we use sentences taken from the first paragraph of each article as it usually contains the summary of the article. We set the number of clusters \texttt{x} to 15,000.

\clearpage
\onecolumn
\subsection{Writing HIT Instructions}
\begin{figure*}[ht!]
    \centering
    \includegraphics[width=\linewidth]{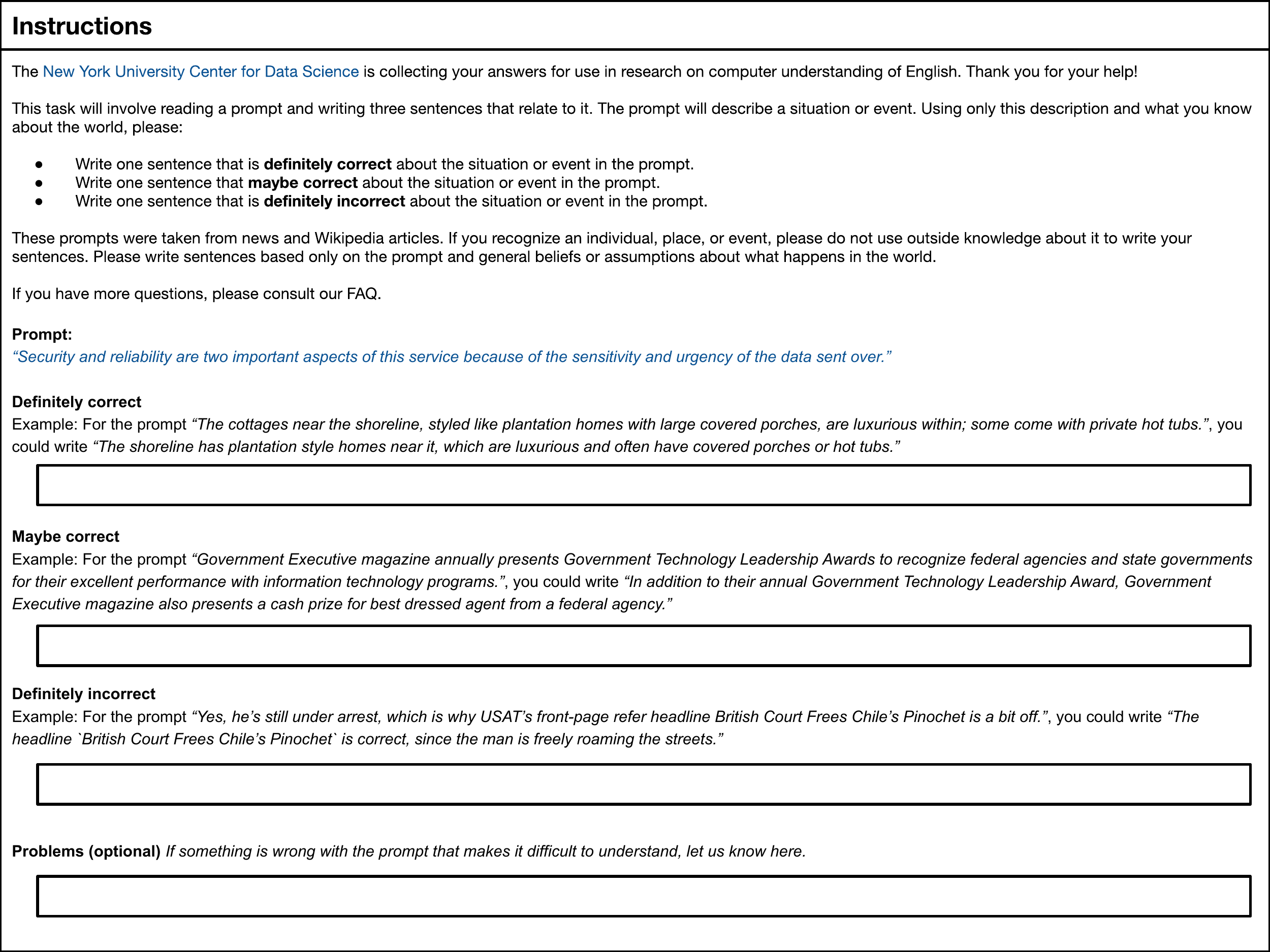}
    \caption{Writing HIT instructions.}
    \label{fig:writing_hit}
\end{figure*}

\clearpage
\subsection{Data Labeling and Validation HIT Instructions}
\begin{figure*}[ht!]
    \centering
    \includegraphics[width=\linewidth]{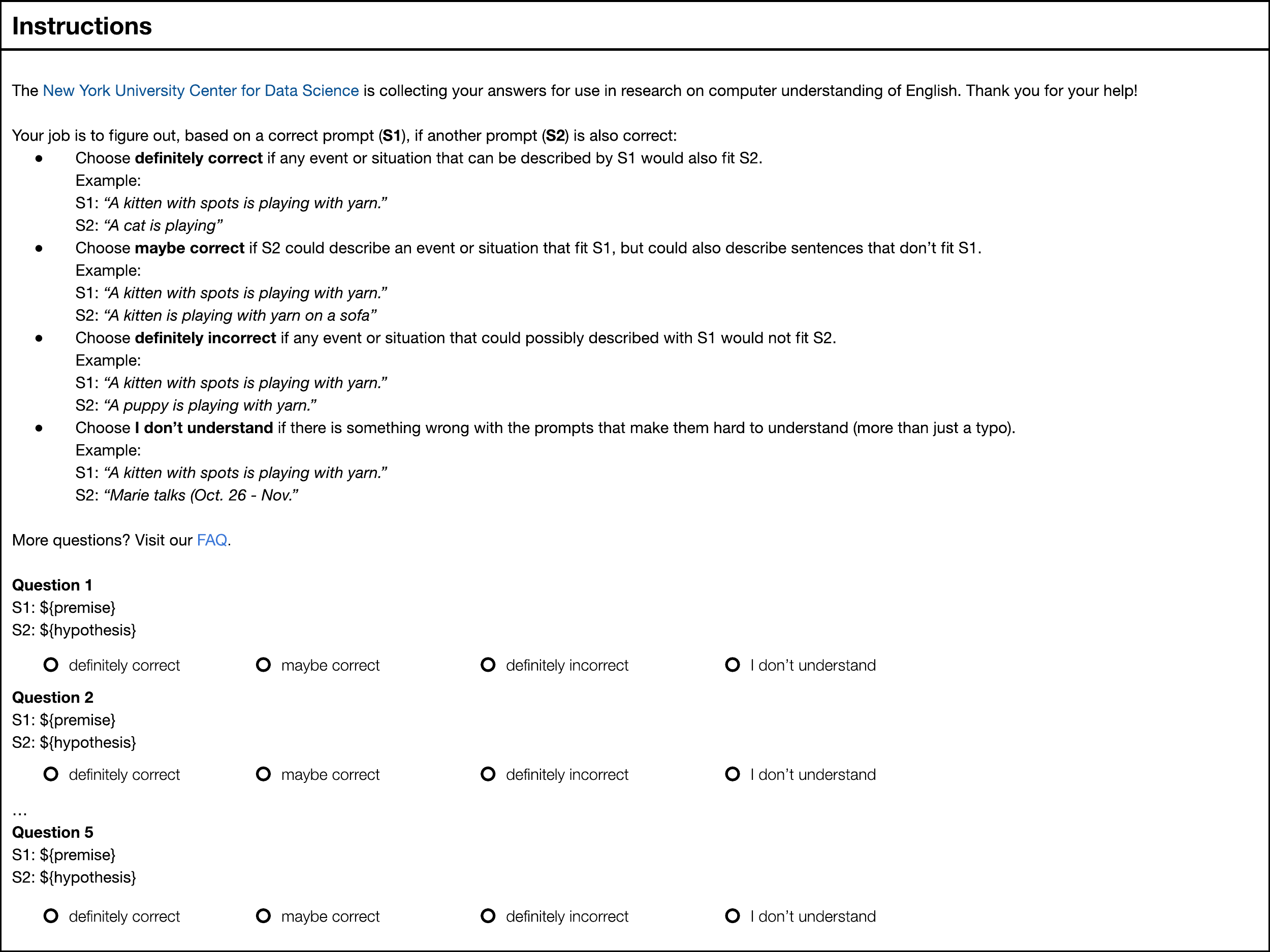}
    \caption{Data Labeling and Validation HIT instructions. We collect one annotation per example for data labeling and five annotations per example for validation.}
    \label{fig:labeling_hit}
\end{figure*}

% \section{Supplemental Material}
% \label{sec:supplemental}

\end{document}